\documentclass{article}

\usepackage[final, nonatbib]{neurips_2023}

\usepackage[utf8]{inputenc} 
\usepackage[T1]{fontenc}    
\usepackage{hyperref}       
\usepackage{url}            
\usepackage{booktabs}       
\usepackage{amsfonts}       
\usepackage{nicefrac}       
\usepackage{microtype}      
\usepackage{xcolor}         
\usepackage[ruled,vlined]{algorithm2e}
\usepackage{amsmath}
\usepackage{graphicx}

\DeclareMathOperator*{\argmax}{arg\,max}

\title{Generation of Games for \\Opponent Model Differentiation}

\author{%
  David Milec \\
  Artificial Intelligence Center\\
  Department of Computer Science \\
  Faculty of Electrical Engineering \\
  Czech Technical University in Prague\\
  \texttt{milecdav@fel.cvut.cz} \\
  \And
  Viliam Lisý\\
  Artificial Intelligence Center\\
  Department of Computer Science \\
  Faculty of Electrical Engineering \\
  Czech Technical University in Prague\\
  \texttt{viliam.lisy@agents.fel.cvut.cz}
  \And
  Christopher Kiekintveld\\
  Computer Science Department, The University of Texas at El Paso\\
  \texttt{cdkiekintveld@utep.edu}
}

\begin{document}

    \maketitle
    
    \begin{abstract}
    Protecting against adversarial attacks is a common multiagent problem. Attackers in the real world are predominantly human actors, and the protection methods often incorporate opponent models to improve the performance when facing humans. Previous results show that modeling human behavior can significantly improve the performance of the algorithms. However, modeling humans correctly is a complex problem, and the models are often simplified and assume humans make mistakes according to some distribution or train parameters for the whole population from which they sample. In this work, we use data gathered by psychologists who identified personality types that increase the likelihood of performing malicious acts. However, in the previous work, the tests on a handmade game could not show strategic differences between the models. We created a novel model that links its parameters to psychological traits. We optimized over parametrized games and created games in which the differences are profound. Our work can help with automatic game generation when we need a game in which some models will behave differently and to identify situations in which the models do not align.
    \end{abstract}
    
    \section{Introduction}
    The real world poses multiple challenging problems in the multiagent security domain. Designing game theoretic algorithms for planning security routes in the airport or for patrol boats near the coastline and other domains are problems already tackled by previous work \cite{an2013deployed, nguyen2016capture, delle2014game, fang2017paws}. Network security is also a topic where multiagent systems are frequently used \cite{ho2022game, amini2023game}. Many previous works search for an optimal strategy when the opponent acts rationally \cite{Pita08, jain2010software}. However, in the real world, the main attacker is often human or some automatic system with limited rationality. Therefore, it is desirable to model the opponent, and many works show that modeling the opponent significantly improves performance against humans \cite{yang2011improving, yang2012computing, nguyen2013analyzing}.

    However, modeling humans is a complex problem, and the models often attempt to model some mistake distribution, for example, quantal response or quantal level-k \cite{yang2012computing}. More complex models add a subjective utility function, which, in conjunction with a quantal response, is used in SUQR and later SHARP and tries to link the model with some basic psychological traits \cite{nguyen2013analyzing, kar2017cloudy}. Furthermore, existing models attempt to model the population as a whole, which improves the performance of the algorithms but has its limits. We follow work proposing to divide humans further and focus on personality types with a higher chance of committing criminal activities \cite{curtis2021dark, forsyth2012meta}. However, in the work, the authors could not demonstrate the strategic differences between the chosen personality types. The results showed that designing a game that would showcase the differences and enable studying the situations in which the personality types behave differently is not trivial. We aim to create such a game automatically.

    To determine if the types are strategically different, we first needed to model them since repeating the human experiment multiple times would be expensive. We took known models and created a new model where the parameters better correspond to the psychological traits of humans so that a researcher can interpret the parameters and even set them manually. We showed that our model performs comparably to the other models on small games. Furthermore, since we base it on reinforcement learning, it is slower on smaller games, but unlike the other models, it can scale very well to larger games.

    Finally, we proposed a method where we train the models to represent the personality types. This can be done by having some data on how the types played or by setting the parameters of the models with the help of psychologists. We had data from the previous experiments and trained the models on the data. Next, we designed a class of games that can be easily parametrized and optimized over the parameters of the games with fixed models. This gives us a general method that can automatically generate games on which we can prove the strategic differences of some personality types and also show where those differences come from.

    \section{Background}
We model the games as factored observation stochastic games (FOSG) \cite{kovavrik2022rethinking}. In FOSG, we have world states in which all players can play, and a combination of actions transfers the players to a new world state using the transition function with some probability distribution. After each step, both players receive a reward. \textbf{Strategy} $\sigma_i$ is a collection of tuples for all world states and observation histories, where each tuple describes how a player $i$ will act in some world state with particular observation history, possibly with some probability distribution over actions. \textbf{Expected utility} is a sum of all the rewards given to both players following some fixed strategy. \textbf{Best response} to a strategy $\sigma_1$ is a strategy that maximizes expected utility against $\sigma_1$ \cite{fudenberg1991game}.

    \section{Model}
In this section, we describe the new model. We based the model on reinforcement learning and chose the distributional q-learning \cite{watkins1992q, bellemare2017distributional}. Q learning uses parameters that nicely correspond to a few human traits. We have $\gamma$, used mostly for convergence in reinforcement learning, but changing $\gamma$ changes the fixed point to which we converge. Since $\gamma$ affects our weight of the future state, it nicely reflects how much the agent plans. By changing $\gamma$, we can effectively create agents with a lookahead of only a few steps or even agents who decide solely on the immediate reward.

Next, we have $\alpha$, which affects how fast the agents learn and often causes problems with convergence. We can use $\alpha$ in the training phase in the same way as in standard q-learning. However, in the test phase, we can have an agent who can try to adapt to the opponent's strategy, which humans tend to do, and we would still do some training with different values of $\alpha$. This would correspond to the adaptability of the human.

Finally, we have parameters that are not part of the original q-learning but correspond to some psychological aspect. First is a different perception of rewards and losses. Humans generally tend to give greater value to losses than to rewards, while psychopaths, on the other hand, tend to disregard the possible losses completely \cite{schmidt2005loss, blair2001somatic}. We introduce $\rho$, which can have a value between 0 and 1, where 0 completely disregards the positive rewards, and 1 completely disregards the negative rewards. If we call positive reward $R$ and negative reward (penalty) $P$, the resulting utility is $U = \rho R + (1 - \rho) P$. This is simplified, and we disregard the so-called framing problem and manually set the reference point by defining a reward and a penalty. In future work, somehow learning the reference point would be beneficial.

Next, we have a rationality parameter $\lambda$, the most common parameter in the other models. In our model, it is used similarly as in quantal response, and after we have the expected utilities for actions, we sample one using a soft-max with the $\lambda$. This allows us to model the mistake-making while still retaining the ability to disregard the soft-max by setting very high lambda and only keeping the other biases.

\begin{algorithm}[H]
\SetAlgoLined
 initialize $Z(s, a), \forall s \in S, \forall a \in \mathcal{A}(s)$ as zero distributions (based on the chosen approximation)\\
 \For{each episode}{
  $S$ = initial state\;
  choose next action A based on $Z$ (for example epsilon greedy/soft-max)\;
  \For{each step in the episode}{
    take action $A$ and observe reward $R$, penalty $P$ and next state $S'$\;
    $Z(S,A) = Z(S,A) + \alpha(\rho R + (1 - \rho) P + \gamma(Z(S',\argmax_a \mathbf{PT}[Z(S',a)]) - Z(S,A)))$\;
    choose next action A based on $Z$ (for example epsilon greedy/soft-max)\;
    $S = S'$\;
  }
 }
 \caption{sub-rational distributional q-learning}
\end{algorithm}

    \section{Method}
This section describes the method used to generate the games. First, we need to train the models. We can either have set parameters by some experts, or we can have data of games played, and if we know which types played which games, we can train the model on that type's part of the data.

The experimental data from previous works contain humans playing against a uniform random strategy in a flip-it game \cite{basak2018initial}. The game has five nodes starting under the defender's control. The game has five rounds. The attacker can attack one node in each round with an option to pass. The defender can defend one of the nodes, and if they both select the same node, it does not change ownership; when they select different nodes, they flip the node they choose to their side. Attacking costs the attacker some constant $C_i$, and after each round, the attacker gets a reward based on all the nodes he controls, summing rewards $R_i$ for each controlled node. The game used had rewards structure $((10, 8), (10,2), (4,2),(4,8),(10,5))$ where each tuple is a node in the format $(R_i, C_i)$. Data comes from 155 participants who filled the short dark triad (SD3) \cite{jones2014introducing} to evaluate their types. We then cluster those with high scores for each group and train our models on the clusters.

We explored different metrics and, for our purpose, chose a metric that should create games where we can play well and differently against the models. The chosen metric is an expected utility against a best response. We also use normalization of the rewards to focus on the structural properties of the games.

The game class we optimized is a more complex version of the flip-it game, and we enhanced it by playing on a graph. We have N nodes, and the attacker starts controlling only the first node. The game is played through rounds, and both players simultaneously choose an action in each round. The attacker can only attack neighboring nodes of any node he controls and pay the cost based not on the target node but on the edge to increase the effect of strategic planning. The defender can defend any single node, and then the game is the same, and the attacker receives a reward for each node she controls after each round. The parameters we train over are all the costs of the edges. We also have a threshold parameter, and the selected edge is not present in the graph when the cost is under the threshold. Finally, the rewards for the nodes are also parameters.

Finally, we selected the metric using the difference in the expected utility of the best response against the model. We ran hyperparameter optimization over the game's parameters, particularly Population-Based Bandits \cite{parker2020provably}, and took the game with the highest metric. We trained for 2000 episodes and fixed $\alpha = 10^{-3}$ since we do not use adaptive learning yet.

    \section{Experiments}
In this section, we show the results of both the model and its performance and the results on games that significantly increase the differences in behavior between the models.

\subsection{Model Results}
\begin{figure}
    \centering
    \includegraphics[width=\linewidth]{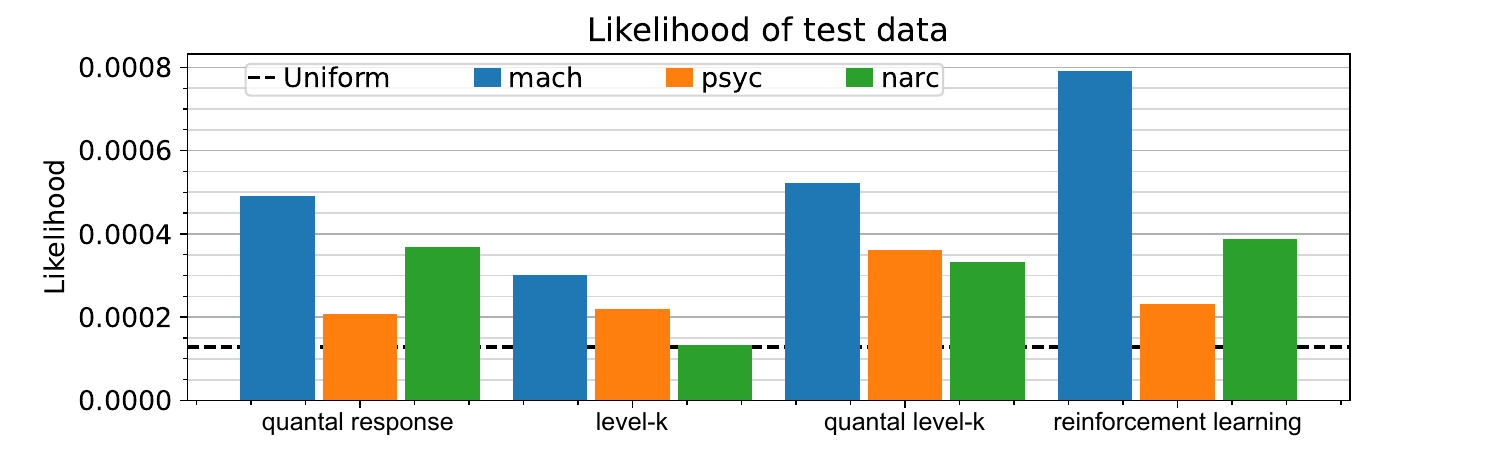}    
    \caption{Test likelihood of the models for dark triad personality types (Psychopaths, Machiavellians, and Narcissists) compared with the likelihood of uniform strategy.}
    \label{fig:models}
\end{figure}

We compare the model to quantal response (QR), level-k (LK), and quantal level-k (QLK). We train the models using 400 trials of PB2 and then report testing likelihood in Figure~\ref{fig:models}. We can see that the new proposed model is competitive, has the highest test likelihood for Machiavellians and Narcissists, and is beaten only by QLK for Psychopaths.

\subsection{Game Generation Results}
We created a game with the highest score using the models to test the ability to generate games that distinguish between models. Then, we tested the differences when we swept over the parameters used in the model. We tried the $\gamma$, $\lambda$ and risk-seeking. We compared differences in expected utility against uniform random and expected utility against best response. We tested with $\gamma$ from 0 to 1, always raising by 0.1. We used $\lambda$ from 0.1 to 12.8; each next value was double the previous one. Finally, $\rho$ was between 0 and 1 with 0.2 increments.

Figure~\ref{tab:difference} shows that the differences generalize over a bigger spread in the parameters. It is always significantly better than the original game when we compare the best response utility we optimized. On the other hand, we lose the spread in the utility against uniform, and a possible future direction would be multi-objective optimization over both.

\begin{table}
    \begin{center}    
    \begin{tabular}{|c|c|c|c|c|c|c|c|c|}
            \hline            
            & \multicolumn{2}{|c|}{\textbf{Original game}} & \multicolumn{2}{|c|}{\textbf{Optimized game}} \\ \hline
            & \textbf{Utility difference} & \textbf{BR utility difference} & \textbf{Utility difference} & \textbf{BR utility difference}  \\ \hline               
            $\gamma$ & 0.102 & 0.105 & 0.139 & 0.473 \\  \hline
            $\lambda$ & 0.347 & 0.266 & 0.429 & 0.456 \\  \hline
            $\rho$ & 0.323 & 0.186 & 0.230 & 0.870 \\  \hline            
    \end{tabular}
    \caption{Differences between the minimal and maximal metric for a sweep over each trait in the original game \cite{curtis2021dark} and our optimized game.}
    \label{tab:difference}
    \end{center}
\end{table}

    \section{Conclusion}
Modeling opponents when facing humans in adversarial attacks is an important part of the solutions, and previous works show a significant increase in performance. However, most of the previous work models humans as a whole when significant psychological research suggests significant differences between personality types. We created a new model to model different human personality types, and the model better corresponds to psychological traits. We trained the model on data from humans. We proposed a method to generate games that show differences between the models and where they behave differently and can then be used in full-scale human experiments. We want to promote this direction in research as it can be a promising direction to further increase the performance of the algorithms in the real world when facing adversarial human agents. In future work, we plan to test our generated games on a full human experiment and look further into the framing problem for the risk aversion parameter of the model.

    \section{Acknowledgements}
    This research was supported by Czech Science Foundation grant no. GA22-26655S and by the Grant Agency of the Czech Technical University in Prague, grant No. SGS22/168/OHK3/3T/13. This work was partially supported by ARO MURI award W911NF1710370. Computational resources were supplied by the OP VVV funded project CZ.02.1.01/0.0/0.0/16\_019/0000765 ``Research Center for Informatics''.

    \bibliographystyle{abbrv}
    \bibliography{bibliography}

\begin{thebibliography}{10}

\bibitem{amini2023game}
M.~Amini and Z.~Bozorgasl.
\newblock A game theory method to cyber-threat information sharing in cloud computing technology.
\newblock {\em International Journal of Computer Science and Engineering Research}, 11(4-2023), 2023.

\bibitem{an2013deployed}
B.~An, F.~Ord{\'o}{\~n}ez, M.~Tambe, E.~Shieh, R.~Yang, C.~Baldwin, J.~DiRenzo~III, K.~Moretti, B.~Maule, and G.~Meyer.
\newblock A deployed quantal response-based patrol planning system for the us coast guard.
\newblock {\em Interfaces}, 43(5):400--420, 2013.

\bibitem{basak2018initial}
A.~Basak, J.~{\v{C}}erný, M.~Gutierrez, S.~Curtis, C.~Kamhoua, D.~Jones, B.~Bo{\v{s}}anský, and C.~Kiekintveld.
\newblock An initial study of targeted personality models in the flipit game.
\newblock In {\em International Conference on Decision and Game Theory for Security}, pages 623--636. Springer, 2018.

\bibitem{bellemare2017distributional}
M.~G. Bellemare, W.~Dabney, and R.~Munos.
\newblock A distributional perspective on reinforcement learning.
\newblock In {\em International conference on machine learning}, pages 449--458. PMLR, 2017.

\bibitem{blair2001somatic}
R.~J.~R. Blair, E.~Colledge, and D.~Mitchell.
\newblock Somatic markers and response reversal: Is there orbitofrontal cortex dysfunction in boys with psychopathic tendencies?
\newblock {\em Journal of abnormal child psychology}, 29:499--511, 2001.

\bibitem{curtis2021dark}
S.~R. Curtis, A.~Basak, J.~R. Carre, B.~Bo{\v{s}}ansk{\`y}, J.~{\v{C}}ern{\`y}, N.~Ben-Asher, M.~Gutierrez, D.~N. Jones, and C.~Kiekintveld.
\newblock The dark triad and strategic resource control in a competitive computer game.
\newblock {\em Personality and Individual Differences}, 168:110343, 2021.

\bibitem{delle2014game}
F.~M. Delle~Fave, A.~X. Jiang, Z.~Yin, C.~Zhang, M.~Tambe, S.~Kraus, and J.~P. Sullivan.
\newblock Game-theoretic patrolling with dynamic execution uncertainty and a case study on a real transit system.
\newblock {\em Journal of Artificial Intelligence Research}, 50:321--367, 2014.

\bibitem{fang2017paws}
F.~Fang, T.~H. Nguyen, R.~Pickles, W.~Y. Lam, G.~R. Clements, B.~An, A.~Singh, B.~C. Schwedock, M.~Tambe, and A.~Lemieux.
\newblock Paws-a deployed game-theoretic application to combat poaching.
\newblock {\em AI Magazine}, 38(1):23--36, 2017.

\bibitem{forsyth2012meta}
D.~R. Forsyth, G.~C. Banks, M.~A. McDaniel, et~al.
\newblock A meta-analysis of the dark triad and work behavior: a social exchange perspective.
\newblock {\em Journal of applied psychology}, 97(3):557, 2012.

\bibitem{fudenberg1991game}
D.~Fudenberg and J.~Tirole.
\newblock {\em Game theory}.
\newblock MIT press, 1991.

\bibitem{ho2022game}
E.~Ho, A.~Rajagopalan, A.~Skvortsov, S.~Arulampalam, and M.~Piraveenan.
\newblock Game theory in defence applications: A review.
\newblock {\em Sensors}, 22(3):1032, 2022.

\bibitem{jain2010software}
M.~Jain, J.~Tsai, J.~Pita, C.~Kiekintveld, S.~Rathi, M.~Tambe, and F.~Ord{\'o}nez.
\newblock Software assistants for randomized patrol planning for the lax airport police and the federal air marshal service.
\newblock {\em Interfaces}, 40(4):267--290, 2010.

\bibitem{jones2014introducing}
D.~N. Jones and D.~L. Paulhus.
\newblock Introducing the short dark triad (sd3) a brief measure of dark personality traits.
\newblock {\em Assessment}, 21(1):28--41, 2014.

\bibitem{kar2017cloudy}
D.~Kar, B.~Ford, S.~Gholami, F.~Fang, A.~Plumptre, M.~Tambe, M.~Driciru, F.~Wanyama, A.~Rwetsiba, M.~Nsubaga, et~al.
\newblock Cloudy with a chance of poaching: Adversary behavior modeling and forecasting with real-world poaching data.
\newblock International Conference on Autonomous Agents and Multiagent Systems, 2017.

\bibitem{kovavrik2022rethinking}
V.~Kova{\v{r}}{\'\i}k, M.~Schmid, N.~Burch, M.~Bowling, and V.~Lis{\`y}.
\newblock Rethinking formal models of partially observable multiagent decision making.
\newblock {\em Artificial Intelligence}, 303:103645, 2022.

\bibitem{nguyen2016capture}
T.~H. Nguyen, A.~Sinha, S.~Gholami, A.~Plumptre, L.~Joppa, M.~Tambe, M.~Driciru, F.~Wanyama, A.~Rwetsiba, R.~Critchlow, et~al.
\newblock Capture: A new predictive anti-poaching tool for wildlife protection.
\newblock In {\em Proceedings of the 2016 International Conference on Autonomous Agents \& Multiagent Systems}, pages 767--775, 2016.

\bibitem{nguyen2013analyzing}
T.~H. Nguyen, R.~Yang, A.~Azaria, S.~Kraus, and M.~Tambe.
\newblock Analyzing the effectiveness of adversary modeling in security games.
\newblock In {\em Twenty-Seventh AAAI Conference on Artificial Intelligence}, 2013.

\bibitem{parker2020provably}
J.~Parker-Holder, V.~Nguyen, and S.~J. Roberts.
\newblock Provably efficient online hyperparameter optimization with population-based bandits.
\newblock {\em Advances in neural information processing systems}, 33:17200--17211, 2020.

\bibitem{Pita08}
J.~Pita, M.~Jain, J.~Marecki, F.~Ord\'{o}\~{n}ez, C.~Portway, M.~Tambe, C.~Western, P.~Paruchuri, and S.~Kraus.
\newblock {Deployed ARMOR protection: The application of a game theoretic model for security at the Los Angeles International Airport}.
\newblock In {\em Proceedings of the 7th International Joint Conference on Autonomous Agents and Multiagent Systems}, pages 125--132, 2008.

\bibitem{schmidt2005loss}
U.~Schmidt and H.~Zank.
\newblock What is loss aversion?
\newblock {\em Journal of risk and uncertainty}, 30:157--167, 2005.

\bibitem{watkins1992q}
C.~J. Watkins and P.~Dayan.
\newblock Q-learning.
\newblock {\em Machine learning}, 8:279--292, 1992.

\bibitem{yang2011improving}
R.~Yang, C.~Kiekintveld, F.~Ordonez, M.~Tambe, and R.~John.
\newblock Improving resource allocation strategy against human adversaries in security games.
\newblock In {\em Twenty-Second International Joint Conference on Artificial Intelligence}, 2011.

\bibitem{yang2012computing}
R.~Yang, F.~Ordonez, and M.~Tambe.
\newblock Computing optimal strategy against quantal response in security games.
\newblock In {\em Proceedings of the 11th International Conference on Autonomous Agents and Multiagent Systems-Volume 2}, pages 847--854, 2012.

\end{thebibliography}

\end{document}